\def\year{2018}\relax
\documentclass[letterpaper]{article} 
\usepackage{aaai18}  
\usepackage{times}  
\usepackage{helvet}  
\usepackage{courier}  
\usepackage{url}  
\usepackage{graphicx}  

\usepackage{array}
\newcolumntype{L}[1]{>{\centering\arraybackslash}m{#1}}

\usepackage{multirow}
\usepackage{threeparttable}
\usepackage{gensymb}
\usepackage{dblfloatfix}
\usepackage{morefloats}
\usepackage{caption}
\nocopyright

\begin{document}
%
\title{Improving Trajectory Optimization using a Roadmap Framework}
\author{Siyu Dai, Matthew Orton, Shawn Schaffert, Andreas Hofmann, Brian Williams \\
Massachusetts Institute of Technology\\
77 Massachusetts Avenue\\
Cambridge, MA 02139
}
\maketitle
\begin{abstract}
We present an evaluation of several representative sampling-based and optimization-based motion planners, and then introduce an integrated motion planning system which incorporates recent advances in trajectory optimization into a sparse roadmap framework. Through experiments in 4 common application scenarios with 5000 test cases each, we show that optimization-based or sampling-based planners alone are not effective for realistic problems where fast planning times are required. To the best of our knowledge, this is the first work that presents such a systematic and comprehensive evaluation of state-of-the-art motion planners, which are based on a significant amount of experiments. We then combine different stand-alone planners with trajectory optimization. The results show that the combination of our sparse roadmap and trajectory optimization provides superior performance over other standard sampling-based planners' combinations. By using a multi-query roadmap instead of generating completely new trajectories for each planning problem, our approach allows for extensions such as persistent control policy information associated with a trajectory across planning problems. Also, the sub-optimality resulting from the sparsity of roadmap, as well as the unexpected disturbances from the environment, can both be overcome by the real-time trajectory optimization process.
\end{abstract}

\section{Introduction} \label{Intro}

Robotic systems deployed in the real world have to contend with a variety of challenges: light-weight arms or those with series elastic actuators shake when they move, wheels slip, IMUs drift, lidars do not reflect off glass doors, structure light sensors fail outdoors, body-mounted cameras get occluded by appendages, and humans in the environment move quickly and in unpredictable manners. These systems cannot spend an unbounded amount of time searching for an optimal motion plan -- a plan that will ultimately be invalidated by the next sensor reading, a change in the environment, or a slipping wheel. Instead, a motion planner must find solutions rapidly even at the expense of optimality. A motion planner that operates quickly allows the robot to truly react to new information and to feel interactive to humans. In addition to quick generation, these plans need to account for the system's dynamics, be robust to disturbances, and operate faithfully within a higher-level task plan.

The problem of moving a robot safely and efficiently in uncertain environments, however, is a challenging one. Often, there is significant complexity with path planning alone, due to the robot and environment geometry. Coupled with dynamic obstacles and sensor noises, the planning problem only becomes more challenging. Additionally, accounting for dynamics and actuation limits becomes untenable within many frameworks.

Due to the complexity of the overall problem, current motion planning and execution systems do not adequately address all of these challenges simultaneously: they often assume the environment is static, or at least, predictable; many do not simultaneously support collision avoidance and complex dynamics; and many generate completely new trajectories for each planning problem instead of allowing for persistent control policy information associated with a trajectory across planning problems.

We have previously developed \textit{Chekhov}, a reactive motion execution system that addresses these requirements~\cite{hofmann2015reactive}. Chekhov avoids obstacles, incorporates dynamic models and control policies, and observes temporal constraints. However, because Chekhov uses a roadmap approach \cite{kavraki1996probabilistic}, and because robotic motion planning state spaces are typically very large, Chekhov's coverage of the operating workspace is very sparse. As a result, trajectories produced by Chekhov are sub-optimal. In this work, we address this limitation by leveraging recent advances in obstacle-aware trajectory optimization \cite{schulman2014motion}. First, we show that recently developed trajectory optimization techniques, which include some capability to avoid obstacles, are not, by themselves adequate for typical problems. We then show that by formulating trajectory optimization problems based on the Chekhov roadmap, the problems associated with using trajectory optimization alone are solved. Further, we show that the optimized trajectory is superior to (more optimal than) the trajectory produced by the roadmap alone. Thus, the combination results in superior performance in terms of feasibility, optimality, and also planning time. Our future goal is to integrate trajectory optimization into the complete Chekhov motion execution system, so it is essential that the trajectory optimization approach is able to incorporate dynamics and temporal constraints, as well as being able to react quickly to disturbances in planning tasks.

\section{Related Work}

Optimization-based robotic motion planners are attracting more and more attention with the increasing complexity of robots and environments. Covariance Hamiltonian Optimization for Motion Planning (CHOMP)~\cite{ratliff2009chomp}, Stochastic Trajectory Optimization for Motion Planning (STOMP)~\cite{kalakrishnan2011stomp}, Incremental Trajectory Optimization for Real-time Replanning (ITOMP)~\cite{park2012itomp} and TrajOpt~\cite{schulman2013finding} are several state-of-the-art optimization-based planners. In this work, we focus on the TrajOpt planner for three reasons. First, the convex-convex collision checking method used in TrajOpt can take accurate object geometry into consideration, shaping the objective to enhance the ability of getting trajectories out of collision. In contrast, the distance field method used in CHOMP and STOMP consider the collision cost for each exterior point on a robot~\cite{zucker2013chomp}, which means two points might drive the objective in opposite direction. Second, the sequential quadratic programming method used in TrajOpt can better handle deeply infeasible initial trajectories than the commonly used gradient descent method~\cite{schulman2013finding}. Third, customized differential constraints, for example velocity constraints and torque constraints, can be incorporated in TrajOpt. This is an important consideration for Chekhov which aims at building a motion execution system that incorporates system dynamics models and control policies, while respecting additional temporal constraints.

Despite the advantages of optimization-based planners, they are not stand-alone planners and their performance is very sensitive to the quality of initializations. Also, numerical trajectory optimization often suffers from the problem of getting stuck in high-cost local optima. Therefore, a natural thought to improve the performance of optimization-based planners is to combine them with global planners. Some existing work, for example Luna et al.~\shortcite{luna2013anytime} and Campana et al.~\shortcite{campana2015simple}, has proposed online path shortening methods for sampling-based planners. The effect of optimization in those approaches is mostly limited to trajectory smoothing and shortening, and can't account for real-time obstacle avoidance and dynamics constraints. Therefore, those modified sampling-based planners still share the typical slow planning times with other common sampling-based planners. Other researches~\cite{park2015parallel} have presented a combined roadmap and trajectory optimization planning algorithm. However, they additionally focused on avoiding singularities in redundant manipulators and meeting Cartesian constraints resulting in relatively long planning times. In comparison, our approach aims at fast reactive real-time planning in practical planning scenarios, and extensive experiment results in Section \ref{results} show that our approach reaches this goal.

\section{Problem Statement and Approach} \label{approach}



The problem solved by Chekhov is to quickly plan and execute robot motions that accomplish a task specified by a set of temporal and spatial constraints. The inputs to Chekhov can change quickly and unexpectedly with time while the motion is being executed. For practical applications, changes fall into three categories:  1) the current state of the robot changes;  2) the goals to be achieved change;  and 3) an environment obstacle moves in a way that affects the robot. Thus, we define a \textit{disturbance} as such an unexpected change to task goals, environment, or robot state. The system we aim at achieving should react, effectively, instantaneously to disturbances; it should act as if it always, ``instantly'' knows what to do, for any combination of goals and circumstances. This fast reaction is key to providing robots the capability to operate effectively in unstructured, uncertain, fast-changing environments.

We make a number of key assumptions in our approach.  Although these assumptions may seem restrictive, we believe that they are consistent with 
a large class of practical robotic manipulation problems. First, we assume that the manipulation workspace is characterized by a limited set of pre-grasp poses.
Second, we assume that the pre-grasp to grasp motion is short, and is best handled by visual and force servoing loops, rather than open-loop planners.
Third, we assume that the collision environments are not overly complex.
We are not trying to solve ``piano mover'' problems like reaching into tunnels or through a maze of obstacles.
Instead, we assume that there is a small set of potential obstacles, such as a workpiece, a table, another robot, or a human, but that some of these may move.
The emphasis here is on achieving fast performance in typical, practical situations. 

We endeavor to achieve a fast, reactive capability by using a roadmap-based approach. The roadmap represents the static collision-free space, and therefore, is re-used across planning instances. For each pair of nodes in the roadmap, $k$ shortest paths ($k\geq1$) are calculated and stored, so that when dynamic obstacles invalidate some of the edges in the roadmap, the probability of finding a collision-free path for the planning task can be improved as we increase $k$. Our approach features three key innovations from the previous Chekhov. First, as stated in Section \ref{Intro}, we extend the roadmap approach used previously in Chekhov by incorporating recent advances in obstacle-aware trajectory optimization \cite{schulman2014motion} in order to improve solution optimality and fast reaction to disturbances. Our goal here is to consider the entire solution space, rather than the very sparse one provided by the roadmap. Second, we use a set of practically relevant test environments, rather than random ones or ones that are artificially challenging.
To this end, we have developed three new environments that represent typical motion planning scenarios. We have also included a fourth environment developed previously in the motion planning community.
Third, we use semantic information about the environment to help guide the construction of the roadmap to favor inclusion of poses that are known to be useful. Utilizing semantic information includes making a basic distinction between static and dynamic obstacles. It also includes utilizing knowledge of objects in the environment in order to generate pre-grasp poses that will be useful for manipulating them.

\section{Implementation} \label{implementation}



\begin{figure}[t]
      \centering
      
\framebox{\includegraphics[width=0.75\linewidth]{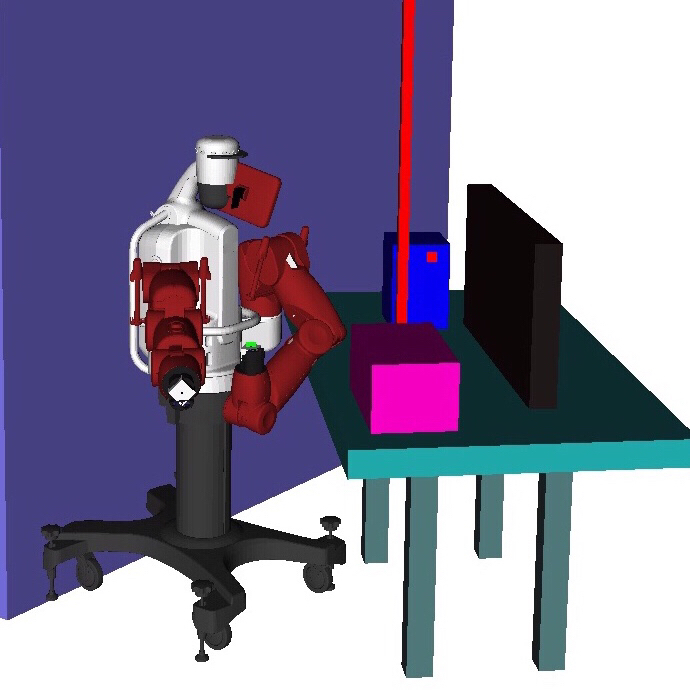}}
      \caption{The ``tabletop with a pole'' environment}
      \label{figure-tabletop}
   \end{figure}

\begin{figure}[t]
      \centering
      \framebox{\includegraphics[width=0.75\linewidth]{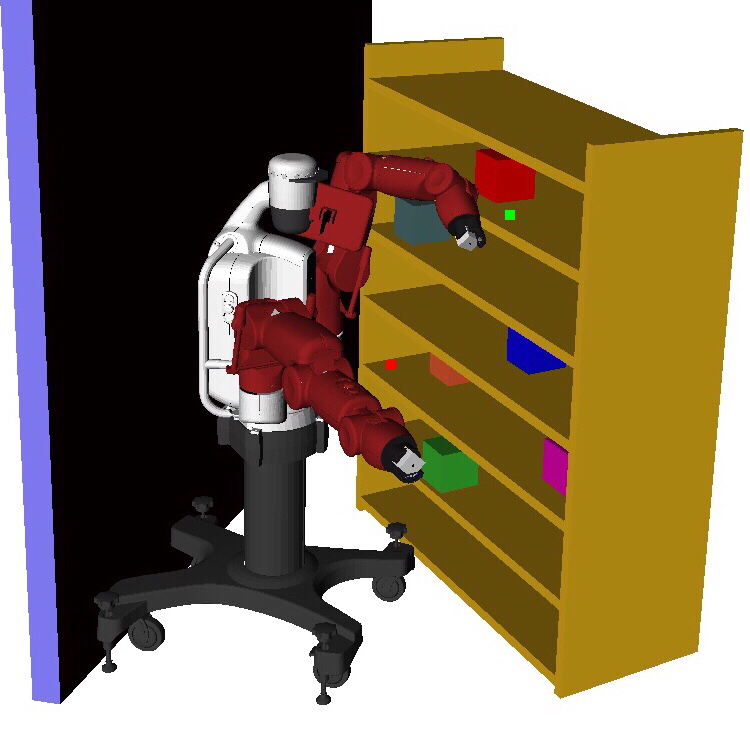}}
      \caption{The ``shelf with boxes'' environment}
      \label{figure-shelf}
   \end{figure}

In order to test and compare the performance of different path planners, we 
use four representational environments: a ``tabletop with a pole'', 
a ``tabletop with a container'', a ``kitchen'' and a ``shelf with boxes''  
environment. We choose environments that are representative of 
different application domains rather than using an environment with randomly-placed obstacles
because our goal is to develop a path planner that operates quickly and 
provides short paths for real world applications. The kitchen environment comes 
from the TrajOpt package, whereas, we designed the remaining three. The 
``tabletop with a pole'' environment, shown in Fig. \ref{figure-tabletop}, is a 
simple tabletop pick-and-place task environment, with a slender pole in the 
middle of the table and a box on each side of the pole. All the planners can 
easily handle most planning queries in this environment. The ``tabletop with a 
container'' environment is similar, but has a large container on the table with 
both boxes inside and outside of it. The ``kitchen'' environment models a 
typical kitchen scenario which is common in household domains. The ``shelf with boxes'' environment, shown 
in Fig. \ref{figure-shelf}, is a 7-level shelf environment with boxes on each 
level of the shelf, which is a common scenario in the logistic application 
domain. This scenario is known to be hard because of the relatively large total number of obstacles and the 
narrow space between them. 

For each environment, we generate 5000 feasible planning tests by randomly sampling 5000 start and target end-effector pose pairs that are collision-free and kinematically feasible. For each sampled point, both the joint-space position and the end-effector location and orientation are recorded. For each experiment trial, planners are provided with the starting joint-space position and the goal end-effector pose. We specify the goal in workspace to give planners the opportunity to find different joint-space solutions to the planning problem. We have ensured that all test cases have a solution by executing all the planners on each test case, and re-sampling start and goal points when no planners could find a solution. All the test cases, including the environment and poses, are saved so that they can easily be repeated in the future.


In our experiments, we use the Baxter robot~\cite{BaxterRobot} with its 7-DOF left arm as the manipulator. Based on our initial tests, TrajOpt works quite similarly on other manipulators, so here we take the left arm as an example to implement the in-depth analysis. 

In addition to the discrete-time collision costs approach, the 
TrajOpt algorithm also provides a ``swept-out volume'' method in order to 
ensure continuous-time collision checking~\cite{schulman2013finding}. 
However, during our experiments, we find that even when the continuous-time collision 
cost is utilized, collision can still occur in-between waypoints, and it is not obvious 
how to use TrajOpt's reported collision cost to detect collisions consistently 
since large cost values can indicate either a collision or just a waypoint close 
to an obstacle. Hence, rather than simply referring to cost values returned by TrajOpt, in our experiments we also 
implement an independent collision checking process for the returned trajectory 
to test continuous-time safety. In particular, we interpolate 100 intermediate 
waypoints between each pair of adjacent waypoints and collision check each point using the 
OpenRAVE collision checking. For our work, we consider this fine-grained discrete-time collision check to approximate a continuous-time collision check sufficiently well.

As introduced previously, we can use semantic information about the environment to 
improve roadmap construction and thus, the motion planning result.
\textit{Semantic Object Maps} (SOM)~\cite{pangercic2012semantic} provide a representation
of such information, including an overall ontology, and also part composition and articulation.
This can be used, for example, to represent how a refrigerator door, or a desk drawer opens and closes, which can be used to generate precise pre-grasp poses for the open and closed positions.
This is important as it guarantees that required poses will exist directly in the roadmap.
Additionally, semantic information can be used to bias sampling of poses during roadmap construction to favor areas of interest.
For example, the area above a desktop is more likely to contain objects of interest and hence should get more nodes than the (free) area under the desktop.

Although our tube-based roadmap architecture supports dynamics and temporal constraints \cite{hofmannwilliams2017}, our experiments here mainly focus on kinematic planning tasks for robot manipulation considering obstacle avoidance. We have already incorporated customized constraints into TrajOpt which respect system dynamics such as torque constraints, velocity limits and acceleration limits. Experiments on planning tasks with dynamics and temporal constraints are beyond the focus of this paper but will be further explored in our future research. Furthermore, for the purposes of evaluating key aspects of our approach, we have assumed that all obstacles in the test environments are static. We focus here on static rather than dynamic obstacles because static obstacles occupy the majority of the workspace in many practical applications. As stated in Section \ref{approach}, we handle dynamic obstacles through storing redundant roadmap paths and by coupling these paths with fast optimization from TrajOpt. Therefore, experiments with dynamic obstacles can be straightforwardly extended from our current experiments.

\section{Experiments and Results} \label{results}

In Section \ref{limitation}, we provide experiment results and performance evaluation of five standard path planners (OpenRAVE BasicRRT, OMPL LazyPRM \cite{bohlin2000path}, OMPL PRM* \cite{karaman2011sampling}, OMPL RRT* \cite{karaman2011sampling}, and TrajOpt with a straight-line joint-space initialization).In Section \ref{new-seed}, we show the results and evaluation of four combined planners which pass in a sampling-based planner solution as an initial path (or ``seed path'') to TrajOpt. Their performance is analyzed and compared in terms of failure-rate, average joint-space path length and average algorithm runtime. Additionally, we also implemented our own roadmap planner which can provide seed paths to TrajOpt -- the results and evaluation of which is described in Section \ref{roadmap}. Each of the experiments includes 5000 test queries and is conducted in all the four environments mentioned in Section \ref{implementation}, but for brevity, most of the tables only provide the results summary for the ``tabletop with a pole'' environment and the ``shelf with boxes'' environment, which qualitatively represent the easiest and hardest environments for the planners, respectively.

\subsection{Limitation of current planners} \label{limitation}

Currently, popular path planners include sampling-based path planners, which 
can operate stand-alone, and trajectory-optimization type path planners, which 
modify a seed trajectory and return the optimized solution. However, in 
practical application scenarios, each of those planners has their own 
disadvantages. The sampling-based path planners are usually not fast enough for real-time planning tasks, and some of them (like PRM and PRM*) 
can not incorporate dynamic constraints. Meanwhile, trajectory-optimization 
type planners locally optimize a path, thus their performance depends much on the quality of seed trajectories. When provided 
with a bad seed, trajectory-optimization type planners can have high collision-rates or get stuck in local optima. This section provides a systematic empirical study on some sampling-based planners and a trajectory-optimization type planner, TrajOpt~\cite{schulman2013finding}, comparing 
their performance in terms of failure-rate, average joint-space path length, and average algorithm runtime. 

We compared five off-the-shelf planners (OpenRAVE BasicRRT, OMPL LazyPRM, 
OMPL PRM*, OMPL RRT* and TrajOpt with straight-line joint-space initialization) on 
all 5000 cases for each environment. For the sampling-based planners, we set 
the runtime upper bound for generating a plan to 300s. The runtime upper bound was choosen, after initial testing, to reduce the failure rates of the optimal sample-based planners (RRT* and PRM*).
For example, if we set the RRT* runtime bound to 60s, the failure rate for the ``shelf with boxes'' environment will be as high as 70\%. 

TrajOpt works by formulating the kinematic motion planning problem as a non-convex 
optimization problem over a $T \times K$-dimensional vector, where $T$ is the number of time-steps and $K$ is the 
number of degrees of freedom~\cite{schulman2013finding}. Hence every trajectory 
in TrajOpt is made up of $T$ waypoints, where the number $T$ is set by the user. 
We ran 16 sets of tests, each with an increasing total number of waypoints, and observed 
that TrajOpt runtime increased approximately linearly with number of waypoints while the 
collision rate dropped quickly with more waypoints. For our tests on TrajOpt with straight-line seed trajectories, we found that
setting $T=30$ provided a good balance between low collision rates and algorithm runtimes. Henceforth, in this subsection, we use 30 total waypoints (including the start and target waypoints).

Table~\ref{table_planners} summarizes the experiment results in the easiest 
environment, ``tabletop with a pole'', and the hardest environment, ``shelf 
with boxes'', in terms of failure rate, average runtime and average joint-space 
path length. The reported failure rate encompasses all possible failure modality (i.e., not finding a solution or returning a solution in collision).
Since TrajOpt will always return a ``solution'' even if the optimization fails, we log a failure when our (secondary) collision checker determines the solution to be in collision;
for sampling-based planners, failure rate is represented by the percentage of cases where the planner failed to return a solution.

If we compare the failure rate of different planners in Table~\ref{table_planners}, we can see that, both in the relatively easy ``tabletop with a pole'' environment and in the relative hard ``shelf with boxes'' environment, TrajOpt fails more frequently to find collision-free solutions than any other planners. If we compare the four sampling-based planners, it can be observed that all the four planners find collision-free solutions for most of the cases in the simple ``tabletop with a pole'' environment. In contrast, in 
the complicated ``shelf with boxes'' environment, RRT and LazyPRM show relatively better solution-finding performance, whereas the optimal planners RRT* and PRM*, even though provided 300s runtime, still fail frequently. From 
the ``average runtime'' column in Table~\ref{table_planners}, it can be 
observed that the sampling-based planners require too much time for most practical path planning applications.
In the case of the optimal planners (RRT* and PRM*), they take all the given time to approximate the optimal solution, therefore their average runtime is always around 300s. Even for LazyPRM, 
7.32s in the simple environment and 63.85 in the complicated environment is infeasible for real-time reaction to disturbances in planning tasks. In terms of average path length, optimal 
planners have noticeable advantages in finding shorter solutions, especially in 
harder environments. Among the remaining planners, LazyPRM tends to return 
longer solutions, which is reasonable due to the intrinsic mechanism of lazy searching algorithms. TrajOpt performance in path length is comparable to sampling-based planners, especially in relatively easy environments.

In conclusion, although sampling-based planners are good at avoiding collision, 
they often take too long for practical application to find a solution. In 
contrast, TrajOpt shows good performance in terms of runtime, but the high collision-rate makes 
it an unsatisfactory practical planner.

\begin{table}[t]
\small

\centering
\begin{threeparttable}
\begin{tabular}{|L{1.7cm}||L{1.4cm}||L{1.35cm}||L{1.35cm}||L{1.4cm}|}
\hline
Environments & Planners\tnote{1} & Failure Rate\tnote{2} & Average Runtime (s)\tnote{3} & 
Average Path Length (rad)\\
\hline
\multirow{5}{1.5cm}{\centering Tabletop with a Pole}
& RRT & 2.30\% & 17.88 & 0.77 \\ \cline{2-5}
& LazyPRM & 0.22\% & 7.32 & 1.76 \\ \cline{2-5} 
& RRT* & 5.32\% & 300.19 & 0.63 \\ \cline{2-5}
& PRM* & 1.00\% & 300.71 & 0.79 \\ \cline{2-5}
& TrajOpt & 17.38\% & 0.56 & 0.71 \\ \cline{2-5}
\hline
\multirow{5}{1.5cm}{\centering Shelf with Boxes} 
& RRT & 10.00\% & 63.86 & 1.06 \\ \cline{2-5}
& LazyPRM & 16.94\% & 63.85 & 2.08 \\ \cline{2-5}
& RRT* & 26.78\% & 300.37 & 0.93 \\ \cline{2-5}
& PRM* & 24.34\% & 300.79 & 1.16 \\ \cline{2-5}
& TrajOpt & 32.06\% & 1.59 & 1.51 \\ \cline{2-5}
\hline
\end{tabular}
\begin{tablenotes}
 \item[1] For each planner in each environment, 5000 planning tasks are tested and the data shown in this table are averaged from the 5000 results.
 \item[2] For TrajOpt with a straight-line seed, failure rate is the percentage of cases where the solution is in collision; for sampling-based planners, failure rate is the percentage of cases where the sampling-based planner failed to find solution.
 \item[3] The runtime upper-bound is set to 300s. RRT* and PRM* always use the full amount of time -- the small deviation from 300s shown in the table is due to small timing errors during simulation.
 \end{tablenotes}
 \end{threeparttable}
 \caption{Evaluation of Current Sampling-based and Trajectory Optimization Planners}
 \label{table_planners}
\end{table}

\subsection{TrajOpt performance with a collision-free seed} \label{new-seed}

The way TrajOpt works indicates its sensitivity and dependency on the 
initialization condition~\cite{schulman2013finding}. Therefore, we propose that 
the performance of TrajOpt can be dramatically improved if we pass in a 
collision-free trajectory as a seed instead of using the joint-space straight-line 
seed. Based on the sampling-based planner experiment results from Section~\ref{limitation}, we 
conduct systematic tests on TrajOpt's performance when provided with a 
sampling-based planner solution as a seed trajectory. For the cases where a 
sampling-based planner found a solution, we pass in the solution as the 
seed trajectory to TrajOpt and record the TrajOpt runtime, solution path length, 
and collision rate. 

TrajOpt algorithm requires the number of waypoints in the solution trajectory to be 
the same as in the seed. Therefore, if we pass in seeds directly from 
sampling-based planners without any pre-processing, the number of waypoints in 
different cases will fluctuate drastically. As mentioned in Section~\ref{limitation}, TrajOpt runtime increases approximately linearly as the number of waypoints 
increases, which means the variation of waypoint numbers will influence runtime. 
Additionally, seeds taken directly from the sampling-based planners with a fewer number of waypoints will results in higher collision rates after processing by TrajOpt than those with more waypoints.
This is because such cases 
usually have longer edges in-between waypoints and are more likely to have seed 
paths that are very close to obstacles. Our tests show that TrajOpt has a much 
weaker ability to deal with edge collisions than with waypoint collisions, and it 
is likely to push path edges into obstacles when shortening and smoothing the 
trajectory. Hence, before passing the seed paths into TrajOpt, we sample 
them by setting a upper bound of 0.16 rad for the distance between adjacent 
waypoints. This pre-processing dramatically reduced the collision rate of 
TrajOpt solutions, as well as narrowing down the variance of TrajOpt's runtime among 
different cases. Inevitably, the average TrajOpt runtime is increased because of 
more waypoints after sampling the seed, but it is still generally under 1s, which is acceptable for real-time planning tasks. 

The performance of this combined ``seed + TrajOpt'' planner is shown in Table~\ref{table_trajopt-seed}.
Comparing the TrajOpt runtime column in Table~\ref{table_trajopt-seed} and the straight-line seed 
TrajOpt runtime in Table~\ref{table_planners}, we see that when provided 
with a good seed, the TrajOpt runtime generally decreased. Specializing to the cases where TrajOpt with a straight-line seed failed to push the trajectory out 
of collision, we found a 50\% - 70\% runtime drop after provided with sampling-based planners' solutions as initializations. Although a small 
percentage of cases end up in collision when TrajOpt is smoothing and 
optimizing the seeds, if we compare the ``average path length'' column in Table~\ref{table_planners} and Table~\ref{table_trajopt-seed}, an obvious improvement 
in average joint-space path length is observed. After comprehensively 
comparing TrajOpt's performance with a sampling-based planner seed and with a straight-line seed, we see
that TrajOpt's performance improves tremendously in terms of both success 
rate and optimization time when provided with a collision-free seed. 
However, according to the ``average runtime'' for combined planners shown 
in Table~\ref{table_trajopt-seed}, it is not feasible to use sampling-based 
planners as seed planners for practical path planning tasks. Thus, the challenge becomes how to generate a good enough seed quickly.

\begin{table}[t]
\small

\centering
\begin{threeparttable}
\begin{tabular}{|L{1.2cm}||L{1.3cm}||L{1.05cm}||L{1.05cm}||L{1.05cm}||L{1.1cm
}|}
\hline
\multirow{5}{1.2cm}{\centering Environ-ments} & \multirow{5}{1.2cm}{\centering Seed Planners} & \multirow{5}{1cm}{\centering Average TrajOpt Runtime (s)} & \multicolumn{3}{c|}{Seed + TrajOpt Planner} \\ \cline{4-6}
& & & Average Runtime (s)\tnote{1} & Average Path Length (rad) & Collision Rate\tnote{2} \\
\hline
\multirow{4}{1.2cm}{\centering Tabletop with a Pole} & RRT & 0.63 & 18.51 & 0.70 & 1.29\% \\ \cline{2-6}
& LazyPRM & 0.98 & 8.30 & 1.28 & 0.12\% \\ \cline{2-6}
& RRT* & 0.29 & 300.48 & 0.54 & 0.02\% \\ \cline{2-6}
& PRM* & 0.36 & 301.07 & 0.64 & 0.10\% \\ \cline{2-6}
\hline
\multirow{4}{1.2cm}{\centering Shelf with Boxes} & RRT & 0.92 & 64.78 & 0.98 & 4.20\% \\ \cline{2-6}
& LazyPRM & 1.36 & 65.21 & 1.60 & 1.57\% \\ \cline{2-6}
& RRT* & 0.46 & 300.83 & 0.81 & 1.17\% \\ \cline{2-6}
& PRM* & 0.67 & 301.46 & 0.95 & 1.98\% \\ \cline{2-6}
\hline
\end{tabular}
\begin{tablenotes}
 \item[1] Sum of sampling-based seed planner runtime (as shown in Table \ref{table_planners} column 4) and TrajOpt runtime averaged from 5000 test cases.
 \item[2] Continuous-time collision rate.
 \end{tablenotes}
 \end{threeparttable}
 \caption{Performance of the Combined ``Sampling-based Seed + TrajOpt'' Planner}
 \label{table_trajopt-seed}
\end{table}

\subsection{TrajOpt with Standard Sampling-based Planner Seed and Roadmap Seed} \label{roadmap}

The core of the roadmap framework for Chekhov is a simplified PRM variant combined with a cache of all-pair-shortest-paths (APSP) solutions. The roadmaps are constructed by randomly sampling points in joint space until a pre-defined number of collision-free points have been sampled.  The sampling is uniform over the four most proximal joints of the robot, and fixed values are assigned to the remaining joints for all nodes.  This approach is taken to more completely cover the workspace with random samples in joint space.  For the tests in Table III and Table IV, the roadmaps start out with 1000 collision-free nodes.  Then, each node is connected to the $k$ nearest neighbors for which collision-free edges exist.  For the tests below, $k=10$ is used.  The resulting graph is pruned of any nodes and edges disconnected from the largest subgraph.  For the environments tested, no more than five of the 1000 points were disconnected from the main subgraph.  Then an APSP solution set is constructed for the pruned roadmap and stored for rapid shortest path queries.

Table III shows the performance of the roadmap planner for all four tested environments.  The remaining two environments omitted in Table \ref{table_planners} and Table \ref{table_trajopt-seed} are also included to emphasize the difficulty of the ``shelf with boxes'' environment relative to realistic environments.  It makes sense that it is difficult to establish collision-free straight-line connections to randomly sampled points in the roadmap when the environment contains narrow shelves with objects inside them.  That being said, tests were conducted to observe the failure rates of roadmaps in different environments relative to the number of randomly sampled points in the roadmap.  As the number of randomly sampled points increased, we observed significant improvement in how often the roadmap was connected to in all environments, particularly in the ``shelf with boxes'' environment.  This leads us to believe that it will not be difficult to develop more intelligent sampling methods that allow roadmaps to more effectively cover all areas of interest within an environment.

\begin{table}[t]
\small

\centering
\begin{threeparttable}
\begin{tabular}{|L{1.8cm}||L{1.3cm}||L{1.3cm}||L{1.35cm}||L{1.3cm}|}
\hline
Environments\tnote{1} & Failure Rate\tnote{2} & Average Runtime (s) & Average Path Length (rad) & Best Average\tnote{3} (rad)\\
\hline
Tabletop with a Pole & 0.18\% & 0.14 & 1.24 & 0.63 \\ \cline{1-5}
\hline
Tabletop with a Container & 0.76\% & 0.18 & 1.32 & 0.80 \\ \cline{1-5}
\hline
Kitchen & 1.92\% & 0.38 & 1.29 & 0.71 \\ \cline{1-5}
\hline
Shelf with Boxes & 12.06\% & 0.39 & 1.30 & 0.93 \\ \cline{1-5}
\hline
\end{tabular}
\begin{tablenotes}
 \item[1] In each environment, roadmap performance is tested on 5000 planning tasks and the data shown in this table are averaged from the 5000 results.
 \item[2] For these roadmaps, failure occurs when no collision-free straight-line connection was found to an existing point on the roadmap from the start or goal pose of a test case.
 \item[3] Best average is the shortest average path length between all tested sampling-based planners in that environment.  Shown here to provide context for the roadmap performance.
 \end{tablenotes}
 \end{threeparttable}
 \caption{Roadmap Performance in All Environments}
 \label{table_roadmap}
\end{table}

If we compare the results in Table III to those in Table I, we can see that, in terms of failure rate, our roadmap planner performs comparably or better than all tested sampling-based planners.  In the most difficult environment, only RRT was able to produce a solution more often than our roadmap planner.  In addtion to failure rate, our roadmap planner's average runtime is substantially better than the sampling-based planners' in all cases.  It is faster by more than an order of magnitude in most observed cases.  This is a result of caching the APSP solution set for fast queries.  Additionally, it should be noted that the roadmap planner constructs the roadmap for each environment a priori whereas LazyPRM constructs a new roadmap online for each case in our tests.  For path length, the roadmap planner performs worse than the optimal planners and RRT, but better than LazyPRM.  In general with roadmap based planners, the sparsity of the roadmap restricts ability to obtain short paths.  With only 1000 nodes, we consider the roadmaps we are using to be relatively sparse for the workspace.  That being said, the roadmap planner generates direct, collision-free paths compared to the off the shelf sampling-based planners.  Since these paths are just seeds for TrajOpt and their lengths are well within an order of magnitude of one another, the discrepancies in path length are not a concern for us.

%

\begin{figure*}[t]
  \centering
    \framebox{\includegraphics[width=0.75\linewidth]{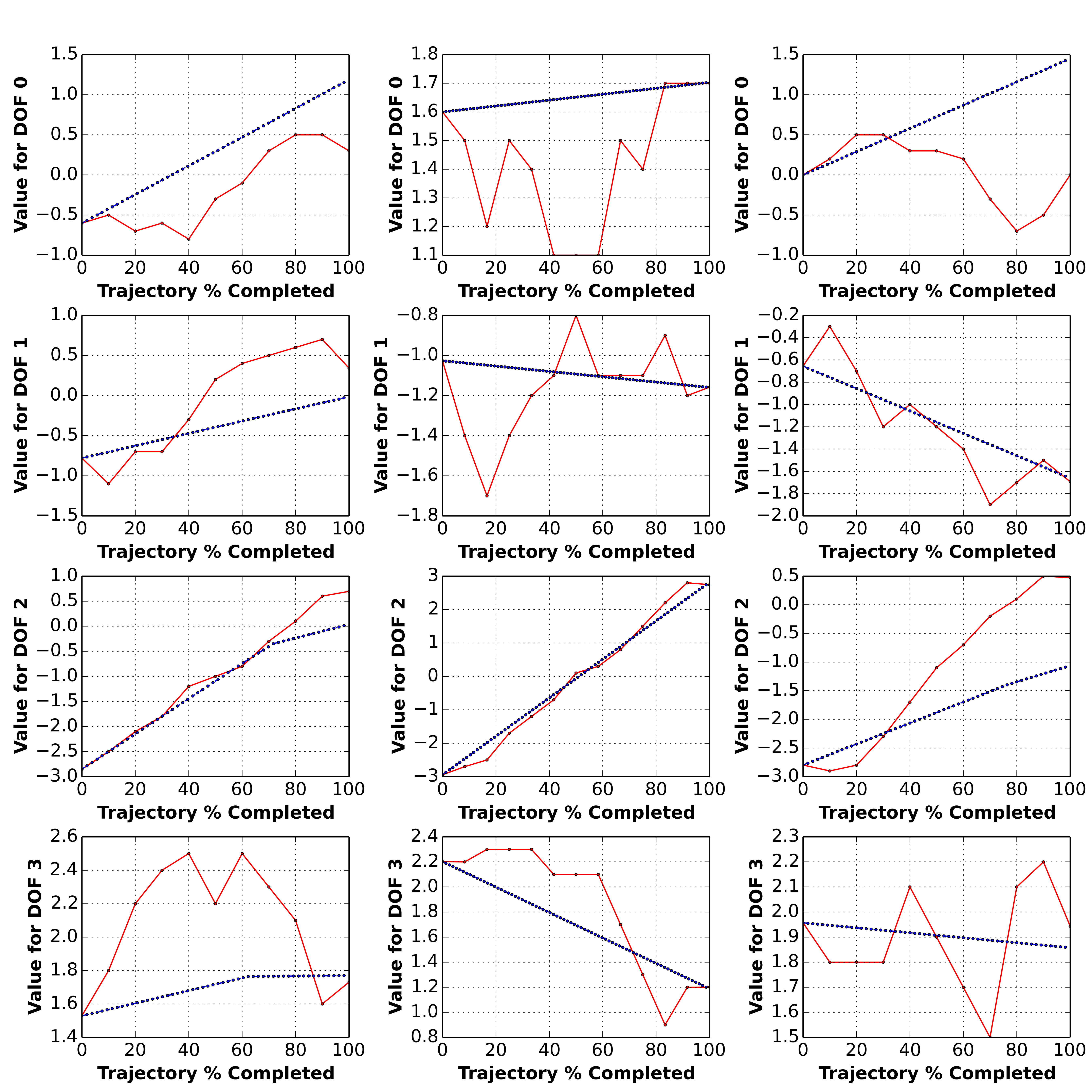}}
    \caption{Roadmap seed trajectories shown with corresponding trajectories optimized by TrajOpt to illustrate improvement on the seed. The solid lines are the roadmap seeds and the dashed lines are the outputted trajectories by TrajOpt when provided those seeds.}
    \label{figure-trajopt}
\end{figure*}

Table IV shows a comparison of solutions produced by TrajOpt when traditional sampling-based planners are used versus our roadmap planner. Many of the observations that can be made from this table reinforce observations made from comparing Table III to Table I. Something new to note is that when the roadmap planner produces a solution, TrajOpt in turn produces a collision-free trajectory more than 98\% of the time. Additionally, these optimized trajectories are on average more than 10\% shorter than their corresponding seed trajectories. Figure \ref{figure-trajopt} shows the four proximal joints for three different trajectories to help visualize the improvments TrajOpt is making on the seed trajectories.  The solid lines are the roadmap seeds and the dashed lines are the outputted trajectories by TrajOpt when provided those seeds. From Figure \ref{figure-trajopt} we can see that TrajOpt fulfilled the task of smoothing and shortening the sub-optimal trajectories produced by the Chekhov roadmap. This result is significant because, as a start, it proves that TrajOpt can effectively optimize the roadmap solutions for kinematic planning problems. Therefore, when we fully incorporate all the dynamics and temporal constraints with TrajOpt, we are optimistic that TrajOpt can also fulfill the task of optimizing trajectories for the whole Chekhov motion and execution framework.

\begin{table}
\small

\centering
\begin{threeparttable}
\begin{tabular}{|L{1.03cm}||L{1.16cm}||L{0.93cm}||L{0.85cm}||L{0.93cm}||L{0.93cm}||L{0.85cm}|}
\hline
\multirow{5}{1cm}{\centering Environ-ments} & \multirow{5}{1cm}{\centering Seed Planners} & \multirow{5}{1cm}{\centering Average TrajOpt Runtime (s)} & \multirow{5}{0.9cm}{\centering Aver-age Seed Length (rad)} & 
\multicolumn{3}{c|}{Seed + TrajOpt Planner} \\ \cline{5-7}
& & & & Average Runtime (s)\tnote{1} & Average Path Length (rad) & Colli-sion Rate\tnote{2} \\
\hline
\multirow{3}{1cm}{\centering Tabletop with a Pole} 
& RRT & 0.63 & 0.77 & 18.51 & 0.70 & 1.29\% \\ \cline{2-7}
& LazyPRM & 0.98 & 1.76 & 8.30 & 1.28 & 0.12\% \\ \cline{2-7}
& RRT* & 0.29 & 0.63 & 300.48 & 0.54 & 0.02\% \\ \cline{2-7}
& Roadmap & 0.45 & 1.24 & 0.59 & 0.82 & 0.06\% \\ \cline{2-7}
\hline
\multirow{3}{1cm}{\centering Shelf with Boxes} 
& RRT & 0.92 & 1.06 & 64.87 & 0.98 & 4.20\% \\ \cline{2-7}
& LazyPRM & 1.36 & 2.08 & 65.21 & 1.60 & 1.57\% \\ \cline{2-7}
& RRT* & 0.46 & 0.93 & 300.83 & 0.81 & 1.17\% \\ \cline{2-7}
& Roadmap & 0.61 & 1.30 & 1.00 & 1.02 & 1.98\% \\ \cline{2-7}
\hline
\end{tabular}
\begin{tablenotes}
 \item[1] Sum of seed planner runtime and TrajOpt runtime averaged from 5000 test cases.
 \item[2] Continuous-time collision rate.
 \end{tablenotes}
 \end{threeparttable}
 \caption{TrajOpt Seeded with Sampling-based Planner Solution compared to Roadmap Solution}
 \label{table_trajopt-roadmap}
\end{table}


The difference in average runtime of the different seed planner coupled with TrajOpt is most notable for highlighting the performance improvements provided by our roadmap planner, but runtime as a metric does not reveal the whole picture for many of these planners.  As noted earlier, the optimal planners like RRT* will always use the full allotted time but may have a good non-optimal solution far sooner than that.  Also, in our test cases, LazyPRM constructs its roadmap online for one time use and then searches for a path in that roadmap.  In general, a PRM does not lend itself to single-query problems.  Our roadmap planner precomputes the roadmap and APSP solutions, but is also essentially a PRM.  It would be interesting to compare the performance of our roadmap planner to faster RRT variants, but it is clear to us that the speed provided by querying precomputed solutions from a PRM of some form outweighs any optimization to be had in online search, especially as system dynamics are factored in.

Overall, our roadmap planner performs as well as if not better than the off the shelf sampling based planners we tested.  The performance metrics used are failure rate, average runtime, and average path length.  Since one of our main goals is to develop a reactive motion execution system that can ``instantly'' replan when disturbances occur, average runtime is where we are most concerned with improvement. Fortunately, average runtime is where we saw the greatest improvement when using our roadmap planner to provide seed solutions rather than using other traditional sampling-based planners. Although we are currently not using dynamic obstacles in our experiments, our average online planning time leaves us optimistic that our planner will be able to handle disturbances in planning tasks with fast reaction.

\section{Discussion} \label{discussion}

Our results show the benefit of extending the Chekhov roadmap approach with the TrajOpt algorithm. The speed of both approaches is preserved, and meanwhile the combination produces more optimal solutions than the roadmap approach alone and with less failure than the TrajOpt approach alone. The average runtime of under 1 sec and the success rate of above 98\% in practical application scenarios show that our approach can handle practical planning tasks with fast reaction.
We are currently distinguishing static from dynamic obstacles
to the extent that the roadmap is constructed to not collide
with the static obstacles in the environment, but dynamic 
obstacles introduced at runtime will likley obstruct nodes and 
edges in the roadmap. Incorporating incremental search algorithms to account for these obstructions 
is an active area of research in our group.  We would also 
like to improve the our ability to connect to our roadmaps 
in difficult environments, but since there are already techniques 
that have been shown to improve roadmap coverage with sparse 
sampling \cite{simeon2000visibility}, we are not currently researching 
new approaches to the problem.

Another active area of research in our group concerns the interaction of dynamics and temporal constraints in integrated motion and task planning problems.
We have previously utilized Chekhov's roadmap framework to incorporate dynamics and temporal constraint 
information \cite{hofmann2015reactive}, \cite{hofmannwilliams2017}, and we plan to extend this work using
recent advances in control theory such as Sum of Squares~\cite{majumdar2017funnel} programming.
This is important for challenging underactuated applications like
underwater mobile manipulators operating in the proximity of reefs, and walking robots.

\bibliographystyle{aaai}
\bibliography{ICAPS_2018}

\end{document}